\documentclass{article}
\usepackage{spconf,amsmath,graphicx}

\usepackage[utf8]{inputenc} % allow utf-8 input
\usepackage[T1]{fontenc}    % use 8-bit T1 fonts
\usepackage{hyperref}       % hyperlinks
\usepackage{url}            % simple URL typesetting
\usepackage{booktabs}       % professional-quality tables
\usepackage{amsfonts}       % blackboard math symbols
\usepackage{nicefrac}       % compact symbols for 1/2, etc.
\usepackage{microtype}      % microtypography

\usepackage{amsmath,graphicx}
\usepackage{times}
\usepackage{amssymb}
\usepackage[utf8]{inputenc}
\usepackage{url}
\usepackage{subfig}
\usepackage{array}
\usepackage{float}
\usepackage{grffile}
\usepackage{rotating}

\begin{document}

%\title{Abusing GANs for Unsupervised Segmentation in Digital Pathology}
\title{Unsupervisedly Training GANs for Segmenting Digital Pathology with Automatically Generated Annotations}
%\titlerunning{Unsupervisedly Training GANs for Segmenting Digital Pathology}

\name{Michael Gadermayr$^1$, Laxmi Gupta$^1$, Barbara M. Klinkhammer$^2$, Peter Boor$^2$, Dorit Merhof$^1$} % <-this % stops a space
%\authorrunning{Michael Gadermayr et al.}

\address{$^1$ Institute of Imaging \& Computer Vision, RWTH Aachen University, Aachen, Germany\\
	$^2$ Institute of Pathology, RWTH Aachen University, 
	University Hospital Aachen, 52074 Aachen,
	Germany}

%\institute{$^1$ Aachen Center for Biomedical Image Analysis, Visualization and Exploration
%			(ACTIVE)\\ Institute of Imaging and Computer Vision, RWTH Aachen University, Germany\\
%			$^2$ Institute of Pathology, University Hospital Aachen, RWTH Aachen University, Germany}

\maketitle

\begin{abstract}	 
	Recently, generative adversarial networks exhibited excellent performances in semi-supervised image analysis scenarios.
	In this paper, we go even further by proposing a fully unsupervised approach for segmentation applications with prior knowledge of the objects' shapes.
	We propose and investigate different strategies to generate simulated label data and perform image-to-image translation between the image and the label domain using an adversarial model.
	%Specifically, we assess the impact of the annotation model's accuracy as well as the effect of simulating additional low-level image features.
	For experimental evaluation, we consider the segmentation of the glomeruli, an application scenario from renal pathology.
	Experiments provide proof of concept and also confirm that the strategy for creating the simulated label data is of particular relevance considering the stability of GAN trainings.
\end{abstract}

\section{Motivation}

% Digital pathology
Due to the progressing dissemination of whole slide scanners generating large amounts of digital histological image data, image analysis in this field has recently gained significant importance~\cite{myHou16a,myBenTaieb16a,Gadermayr18d,myValkonen17a,myVeta16a,Gadermayr17d,myHerve11a}. 
%
%Considered applications mostly consist of either segmentation~\cite{myBenTaieb16a,Gadermayr17d}, classification~\cite{myHou16a,myBarker16a,mySertel09a,myValkonen17a} or regression tasks~\cite{myVeta16a}.
%For segmentation tasks, recently fully-convolutional networks~\cite{myBenTaieb16a,myRonneberger15a} yielded excellent performances in combination with a high computational efficiency compared to sliding-window classification.
%
% Challenges
For segmentation applications, especially fully-convolutional networks proofed to be highly effective tools~\cite{myRonneberger15a,myBenTaieb16a,Gadermayr17d}.
A major challenge in the field of digital pathology is given by a large set of different application scenarios as well as changing underlying data distributions which is due to inter-subject variability, different staining protocols and/or pathological modifications~\cite{Gadermayr18b}.
Each individual application scenario therefore requires large amounts of annotated training data covering the prevalent variability. 
%If the variability is not taken into account, modern machine learning approaches are typically vulnerable to changes in this characteristics between training and testing~\cite{Gadermayr18b,Gadermayr17b}.
The acquisition of such large amounts of labeled training data, however, is typically time-consuming and cost-intensive and thereby constitutes a burden for the deployment of automated segmentation techniques.

% Solutions
%Recently, large effort was taken to train models with a small amount of training data~\cite{Gadermayr17c}. 
For training state-of-the-art machine learning approaches such as fully-convolutional networks, data augmentation proved to be a highly powerful tool~\cite{myRonneberger15a,myRatner17a} to keep the amount of required training data decent. A limitation of data augmentation in combination with supervised learning approaches is given by the fact that often large non-annotated data is available "for free" but is not utilized for training at all. Particularly in the fields of medicine, such as digital pathology, huge amounts of digital image data are routinely captured without any (additional) effort whereby a complete annotation of all data is definitely not feasible. In order to take advantage of non-annotated data as well, dedicated semi-supervised segmentation approaches relying on adversarial models were recently proposed~\cite{myKozinski17a,myIsola16a,myHung18a}.

% I2i-translation: also a solution
Adversarial models were also developed for the field of image-to-image translation~\cite{myJohnson16a,myZhu17a}. Recently, the so-called cycleGAN~\cite{myZhu17a} was introduced which eliminates the restriction of corresponding image pairs for training.
%The authors proposed a generative adversarial network (GAN) relying on a cycle-consistency loss which is combined with an adversarial loss to perform circular trainings, i.e. translations from a domain X to a domain Y and back to domain X and the same vice versa are conducted.
%This GAN architecture not only exhibits excellent performance in image-to-image translation applications.
This architecture can also be utilized for means of unsupervised domain adaptation~\cite{myChartsias17a,myWolterink17a,Gadermayr18d}. 
%For example, images can be converted between MRI and CT in order to finally process (e.g. segment or classify) the images in a (fake) target domain~\cite{myChartsias17a,myWolterink17a}. 
%The huge benefit of this approach is, that it does not require corresponding pairs from both domains which are mostly unavailable or at least difficult to obtain. 
The domain adaptation in these cases is performed on image-level, i.e. "fake" images showing similar characteristics as the target domain samples are generated. This strategy is highly flexible as it can be combined with arbitrary further segmentation or classification approaches.
%is completely transparent (as an observable "fake" image is generated) and consequently can be combined with any segmentation method.
%% TODO cite more work maybe here

\begin{figure}[hb]
	\includegraphics[width=\linewidth]{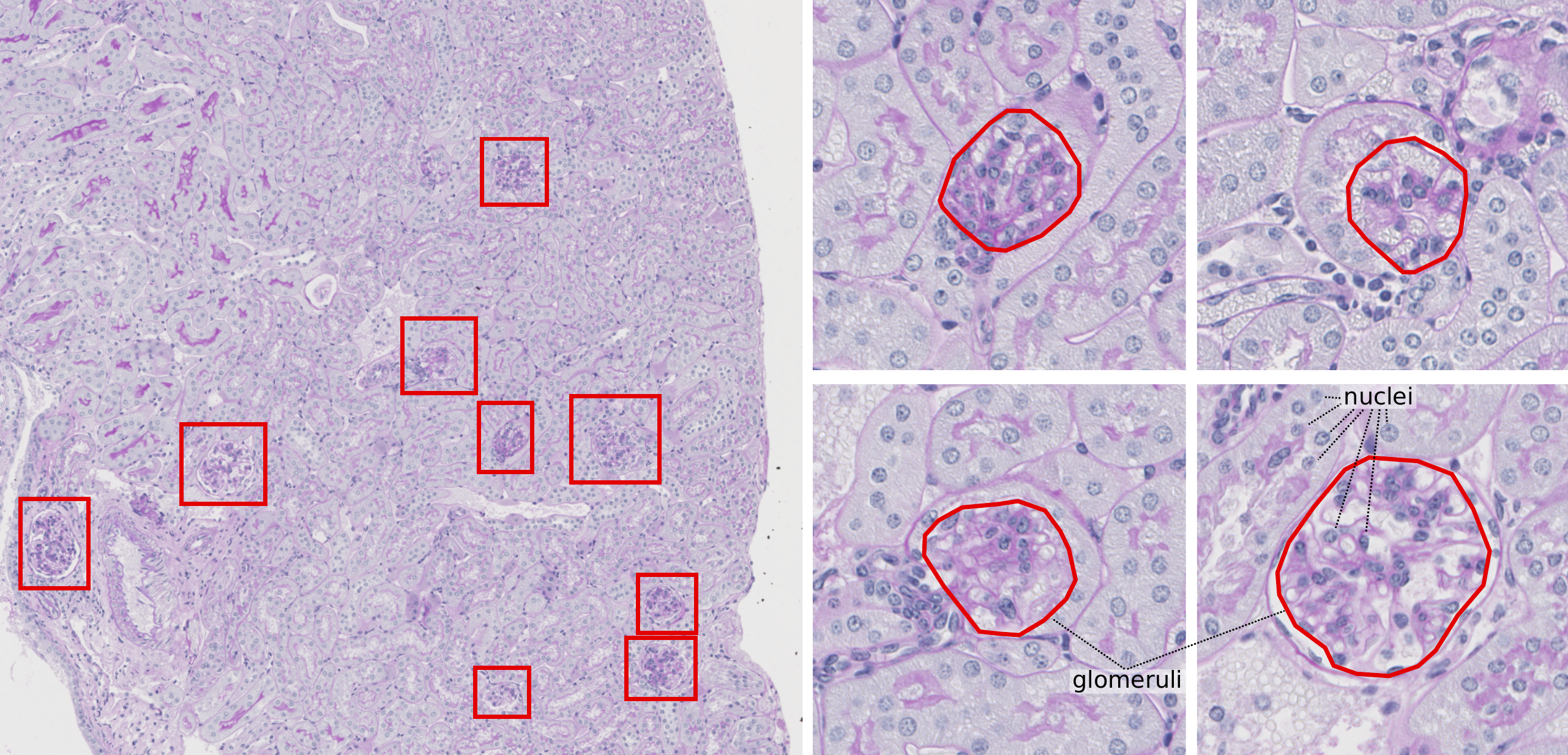}
	\caption{This illustration shows an extract of a renal whole slide image with marked glomeruli (left) as well as magnifications of single glomeruli showing precise manual annotations (right).}
	\label{fig:exampleRenal}
\end{figure}

\noindent \textbf{Contribution:} 
We tackle the problem of acquiring labeled training data by proposing a framework completely bypassing the need for manually labeled objects.
We focus on generating artificial annotations to perform image-to-image translation on unpaired data sets. %based on unpaired data between the image and the label domain.
In our experimental study, we investigate strategies for modeling the shape of the annotations and for modeling additional image information to facilitate training the translation networks.
As application scenario, we consider a segmentation task from digital pathology, specifically the segmentation of the renal glomeruli~\cite{Gadermayr17d,myKato15a,myHerve11a} (Fig.~\ref{fig:exampleRenal}). 
%Although generally showing high variability in texture, color and size, the overall shape of the objects remains almost stable which is a crucial criterion for the proposed method.
%This proof-of-concept provides evidence for the applicability in digital pathology. The evaluation of different settings offers incentive for an investigation with respect to further segmentation tasks in (biomedical) image analysis.

\section{Methods}

For the proposed method, we make use of an image-to-image translation approach. Specifically, we utilize a generative adversarial network (GAN) which facilitates training with unpaired data~\cite{myZhu17a}.
The four subnetworks consisting of two generators and two discriminators are optimized based on an adversarial loss as well as a cycle consistency criterion.
This formulation does not require sample pairs, i.e. there is no need to obtain corresponding image samples for the two domains. Instead it is sufficient to collect a set of images, individually for each domain.
If annotations are interpreted as label (e.g. binary) images, this approach can be utilized for segmentation applications as well. 
%But it was shown that approaches taking the pair-relationship into consideration exhibit better results~\cite{myIsola16a,myZhu17a}.
%However, obtaining such pairs in a segmentation scenario typically means that each original image needs to be annotated such that a pair consists of an original and a label image.
%For unpaired training~\cite{myZhu17a}, however, there is no need to generate annotations corresponding to original images.
The architecture allows to perform training based on a set of images and a (non-corresponding) set of annotations as long as the annotations are realistic (i.e. the distribution matches the underlying distribution of real annotations).
%Therefore, such a setting only needs one data set containing artificially generated annotations and non-corresponding images.

The proposed method relies on an automated generation of realistic annotation images followed by training an image-to-image translation model which is finally able to convert original images to annotations.
The procedure is based on the following assumptions:
%\begin{enumerate}
	%\item 
	(1) we need to understand the underlying distribution of the annotation data and we need to be able to model this distribution (for details, see Sect.~\ref{sec:model}).
	% TODO how do they need to look like?
	%\item 
	(2) The unpaired image-to-image translation approach needs to be effectively applicable to translate between the image and the annotation domain. If a straight-forward translation is not effective, additional information can be added to the annotation domain to enhance the translation process (for details, see Sect.~\ref{sec:enhance}).
%\end{enumerate}

%TODO experiments how good does the data match?
\vspace{-0.1cm}
\subsection{Annotation Model} \label{sec:model}
In the considered application scenario (Fig.~\ref{fig:exampleRenal}), the underlying distribution (assumption~1) of the objects-of-interest is rather basic and can thereby be approximately modeled quite well. The objects-of-interest show roundish shapes which are sparsely distributed over the kidney. For training we consider patches extracted from the whole slide images. We assume that the number of objects per patch can be approximated by a (quantized) Gaussian distribution $G_{\#} \sim \mathcal{N}(\mu_g,\,\sigma_g^{2})$. The objects are uniformly distributed over the patch with one single further assumption that the objects may not overlap. 
For generating the annotations, we investigate two different approaches.
%\begin{itemize}
	%\item 
	Firstly, we consider the objects-of-interest as round objects (\textbf{Circular objects (C):}). The objects' radii $r$ are randomly sampled from a Gaussian distribution $R \sim \mathcal{N}(\mu_r,\,\sigma_r^{2})$.
	%\item 
	In a second configuration, we incorporate the fact that the objects-of-interest often show an elliptic shape (\textbf{Elliptic objects (E):}). 
	To incorporate this knowledge, $r_1$ is drawn from the same distribution as $r$ and $r_2 = r_1 + r_\delta$ where $r_\delta$ models the eccentricity and is drawn from $R_{\delta} \sim \mathcal{N}(0,\,\sigma_e^{2})$.
	 A further rotation parameter $\alpha$ is drawn from a uniform distribution in the interval $ [ 0, 2 \pi ] $.
%\end{itemize}

\subsection{Image-to-Label Translation} \label{sec:enhance}
The straightforward approach consists of adding either circles or ellipses as binary objects into two dimensional matrices which are interpreted as single channel images.
However, for training the image-to-image translation approach, this setting can be highly challenging due to the loss criteria:

% I2I Translation
For training the GAN~\cite{myZhu17a}, two generative models, $F:\mathcal{X} \rightarrow \mathcal{Y}$ and $G:\mathcal{Y} \rightarrow \mathcal{X}$ and two discriminators $D_X$ and $D_Y$ are trained optimizing the cycle consistency loss $\mathcal{L}_{c}$ 
\begin{equation}\begin{split}
\mathcal{L}_{c}=\mathbb{E}_{x \sim p_{data}(x)} [|| G(F({x})) - {x} ||_1 ] + \\
\mathbb{E}_{y \sim p_{data}(y)} [ ||F(G({y})) - {y}||_1 ]
\end{split}
\end{equation}
and the adversarial loss $\mathcal{L}_{d}$
%as well as the adversarial loss 
\begin{equation}\begin{split}
\mathcal{L}_{d} =
\mathbb{E}_{x \sim p_{data}(x)} [\log (D_X(x)) \hspace{-0.1cm} + \hspace{-0.1cm} \log (1 \hspace{-0.1cm} - \hspace{-0.1cm} D_Y(F(x)))] \;\hspace{-0.1cm}+ \\
\mathbb{E}_{y \sim p_{data}(y)} [\log (1 \hspace{-0.1cm} - \hspace{-0.1cm} D_X(G(y))) \hspace{-0.1cm} + \hspace{-0.1cm} \log (D_Y(y))] \; .
\end{split}\end{equation}
%encouraging indistinguishable fake images based on the domain discriminators $D_X$ and $D_Y$.
$F$ and $G$ try to generate fake images that look similar to
real images, while $D_X$ and $D_Y$ aim to distinguish between
translated and real samples.
The generators aim to minimize this adversarial objective against the discriminators that try
to maximize it.

Let $X$ be the domain referring to the original images and let $Y$ be the label domain.
The cycle criterion requires that an annotation mask can be translated to an image by the generator $G$. The generator $F$, however, hides all low-level image details, such as nuclei and tubuli (Fig.~\ref{fig:exampleRenal}) and only preserves the high-level shapes of the glomeruli. Based on these shapes only, it will not be able to reconstruct e.g. the nuclei at the right (i.e. the same) positions leading to a high cycle-consistency loss even though the images might look realistic.
To take this into account, we propose and investigate a second setting simulating the nuclei exhibiting low-level information as well. As for the glomeruli, the number of nuclei is drawn from a (quantized) normal distribution $N_{\#} \sim \mathcal{N}(\mu_{n},\,\sigma_{n}^{2})$. They are uniformly distributed over the whole patch with the restriction that they may not overlap. Diameter is fixed to $d_n$. The additional binary matrix containing the nuclei is added as further image channel to the annotation image. 
This channel is only needed to train the GAN. For testing, this further channel is simply ignored.
Whereas the setting incorporating only the target labels (i.e. the glomeruli) is referred to as single-class scenario, the setting also incorporating further low-level information is referred to as multi-class scenario. Finally, we identified four different settings: single-class circular objects (SC), single-class elliptical objects (SE), multi-class circular objects (MC) and multi-class elliptical objects (ME). 

To facilitate learning, Gaussian random noise ($\sigma_{f_n}$) is added to the annotation maps followed by the application of a Gaussian filter ($\sigma_{f_s}$) to smooth the objects' borders in all settings.

%\subsection{Image Data}

\subsection{Experimental Setting}

Paraffin sections ($1 \mu m$) are stained with periodic acid-Schiff (PAS) reagent and
counterstained with hematoxylin. Whole slides are digitalized with a Hamamatsu NanoZoomer
2.0HT digital slide scanner and a $20 \times$ objective lens.
From each of the $18$ WSIs overall (6 for each dye), $100$ patches with a size of $500 \times 500$ pixels are randomly extracted. For evaluation purpose, the WSIs are manually annotated under the supervision of a medical expert. For each stain individually, for training and testing, patches from three and three images where used, respectively. % from the cortex of the kidney which contains the glomeruli. %The anatomical structure of the organ allows easy manual delineation of this part because it shows a clear change in texture.
%This part of the kidney can be easily and efficiently manually selected due to a clear change in texture and knowledge of the anatomical structure of the kidney. 
As large context is required to assess whether segmentations are realistic, a (rather low) resolution corresponding to a $2.5 \times$ magnification is utilized (original images downscaled by factor eight). %, i.e. the highest ($20 \times$) resolution images are downscaled by factor $8$.

For image-to-image translation, we make use of the cycle GAN as proposed in~\cite{myZhu17a}.
We rely on the provided pytorch reference implementation. Apart from the following changes, we use the proposed standard settings.
As generator model, a residual network consisting of four blocks is utilized. As discriminator, we rely on the suggested patch-wise CNN with three layers~\cite{myZhu17a}.
Learning rate is fixed to $10^{-6}$, number of training epochs is set to $15$ and batch size is set to one. The losses are equally weighted. For data augmentation, flipping, rotation and random cropping ($384 \times 384$ pixel sub-patches) is performed.

%In the annotation masks, positive pixels are set to 155 and negativ e pixels are set to 100 (instead of 255 and 0 corresponding to min and max in the eight bit images) to facilitate the addition of noise in both directions.
%For training, subimages with a size of $384 \times 384$ pixels are randomly selected in each iteration. 
%The cycle consistency loss and the adversarial loss are equally weighted. No further loss (e.g. identity loss) is used.

The annotations are generated based on the following visually assessed parameters (we did not incorporate statistical information of the data set to avoid introducing significant supervision): 
$\mu_g = 7$,
$\sigma_g = 2$,
$\mu_r = 18$,
$\sigma_r = 2$,
$\sigma_{e} = 2$,
$d_n = 4$,
$\mu_{n} = 5000$,
$\sigma_{n} = 50$,
$\sigma_{f_n} = 5$ and
$\sigma_{f_s} = 2$.
%To avoid introducing a significant amount of supervision, these parameters were obtained by visual assessment only. %of a small subset of the image data. We did not quantitatively analyze the ground-truth annotations (which are available for evaluation purpose) in order and thereby bias to the unsupervised application scenario. 

For evaluation, we investigate two optimization strategies. The first strategy does not incorporate any optimization and we basically report the obtained segmentation performance after training for all 15 epochs. As GAN training is, in general, often unstable, we also optimize the epoch by separating the testing data set into one patch for optimization and the others for testing. We use only one patch for optimization because the approach is intended to be unsupervised. %Nevertheless, the effort required for annotating one patch only is negligible.
%We also report the aggregated testing performance individually after each epoch to assess convergence of the GAN.

Apart from pixel-level scores (F$_1$-score (F), precision (P), recall(R)), we also report the corresponding object-level scores (F$_o$, P$_o$, R$_o$). That means, we distinguish between true positive objects (i.e. the distance between the center of a detected object and a real object is smaller than 10 pixels), objects which were missed and false positively detected objects.

All experiments are repeated four times.
The obtained performances are compared with the a supervised fully-convolutional network~\cite{Gadermayr17d}. 
%We utilize a network which was already applied effectively to the considered application scenario. Specifically, we employ a shallow version of the U-net~\cite{Gadermayr17d,myRonneberger15a} which showed good performances on downscaled WSIs. For details on training and augmentation, we refer to~\cite{Gadermayr17d}. The experiments are conducted on a NVIDIA GeForce GTX 1080 Ti GPU.

\section{Results}

\begin{figure}[tb]
	\centering
	\includegraphics[width=\linewidth]{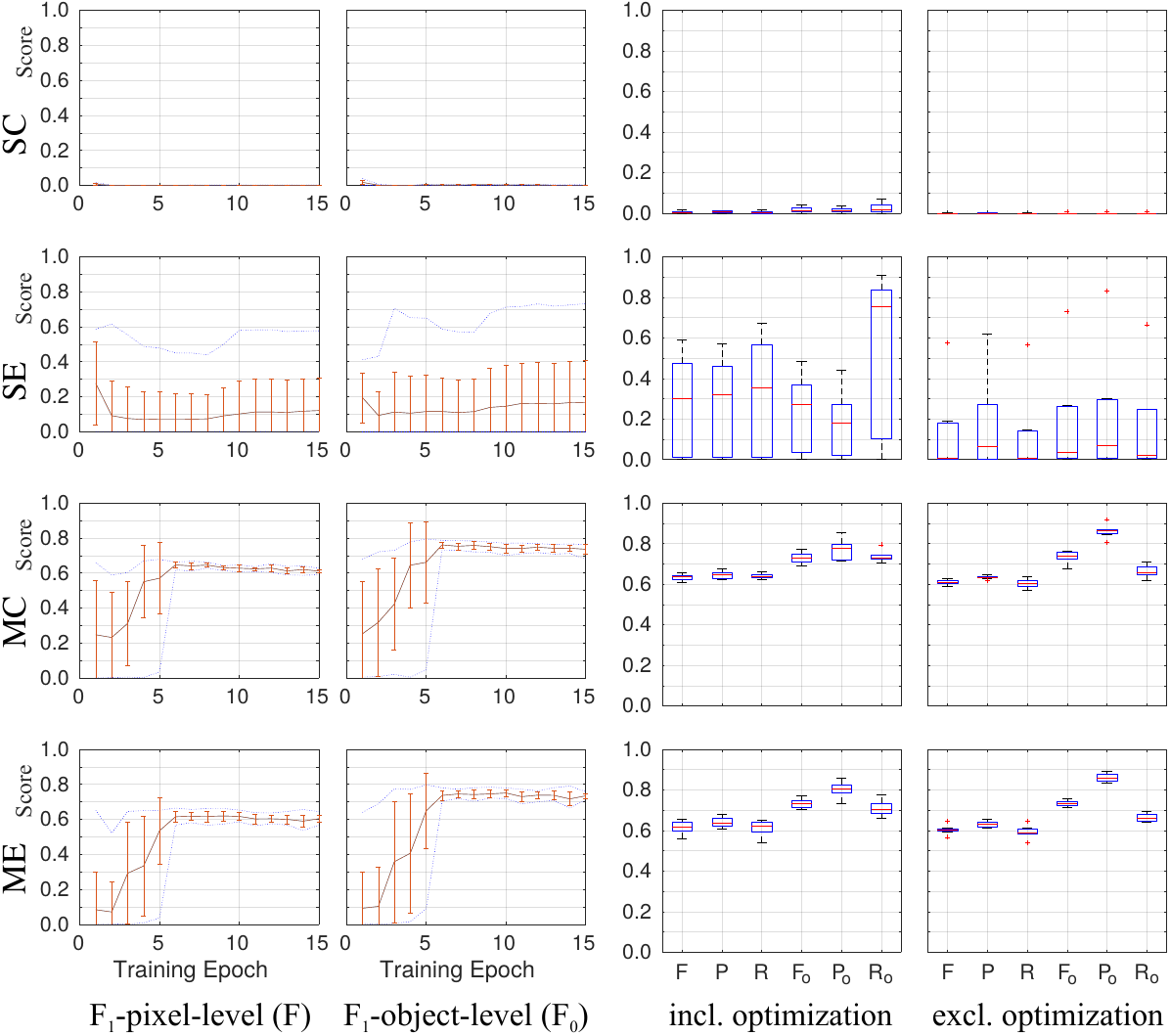}
	\caption{Experimental results for the four different settings (row 1 to row 4). The  left columns show pixel- and object-level F$_1$-scores for testing after varying number of training epochs. The right columns provide F$_1$-scores (F), precision (P) and recall (R) as well as object-level measures (F$_o$, P$_o$, R$_o$) for training for 15 epochs (excl. optimization) and for optimizing the epoch (incl. optimization).}
	\label{fig:res}
\end{figure}

%TODO change subplots (O-F F_O)

\begin{figure}[tb] \center
	\subfloat[F$_1$-pixel-level (F)]{\includegraphics[clip=true, trim=3cm 8.8cm 3cm 8.8cm, width=3.5cm]{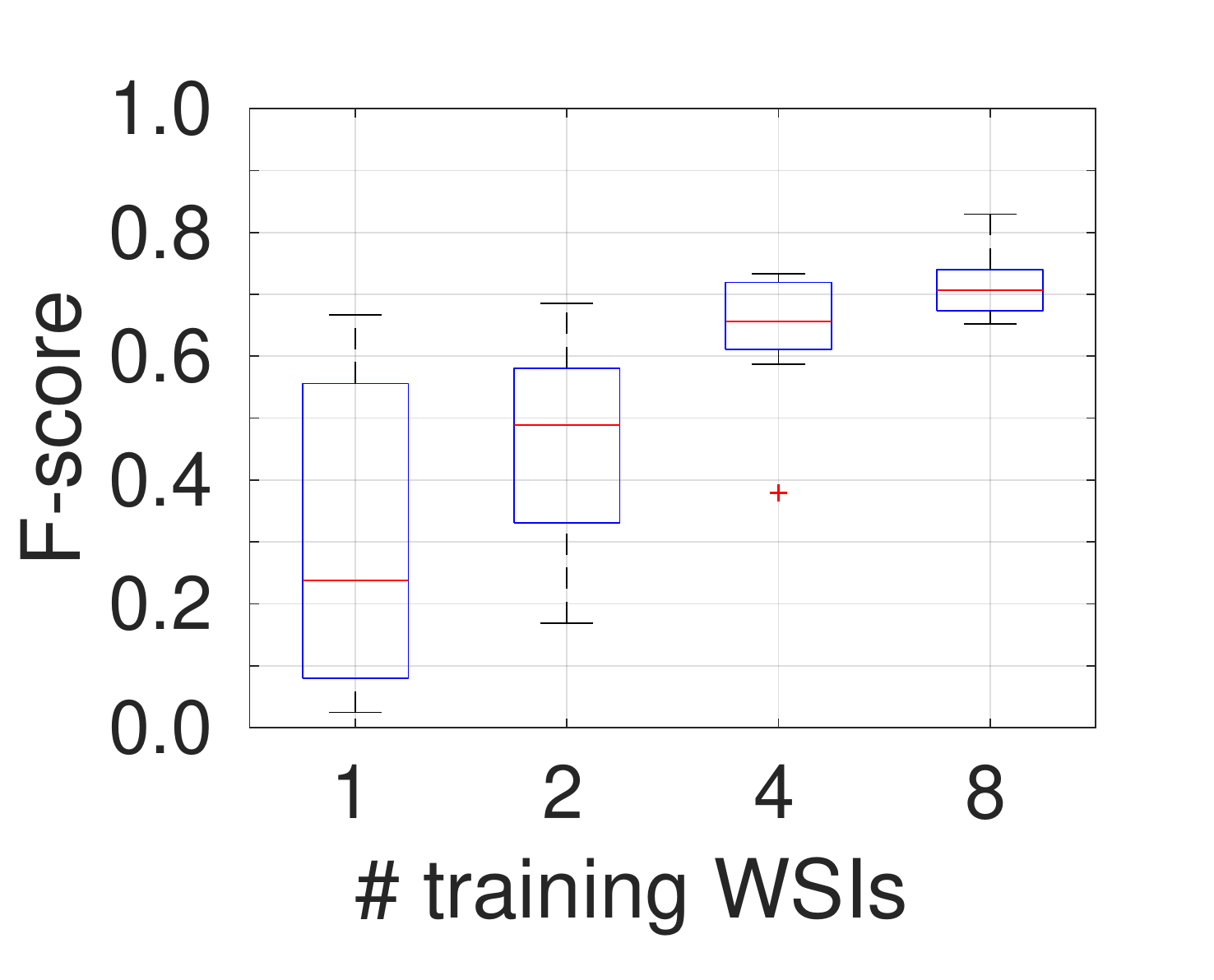}} $\;\;\;\;\;$
	\subfloat[F$_1$-object-level (F$_o$)]{\includegraphics[clip=true, trim=3cm 8.8cm 3cm 8.8cm, width=3.5cm]{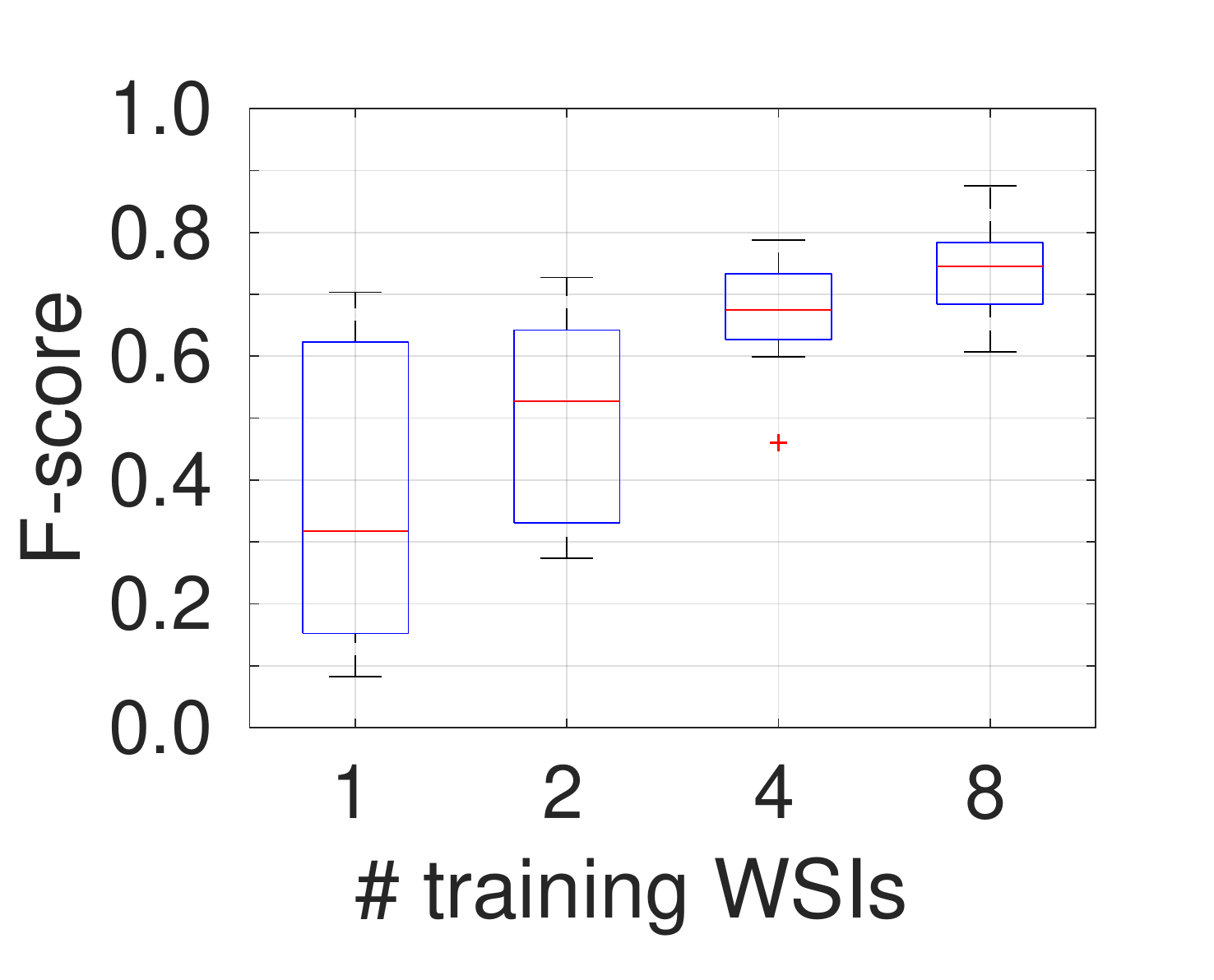}} 
	\caption{Baseline pixel-level (a) and object-level (b) F$_1$-scores indicating the segmentation performance of the supervised U-Net-based approach~\cite{Gadermayr17d} with variable numbers of fully-annotated training WSIs. One single training WSI contains on average 120 single objects.}
	\label{fig:baseline}
\end{figure}

Fig.~\ref{fig:res} shows quantitative results for each of the four different settings. We investigate pixel-level as well as object-level scores. 
The left two columns show the testing pixel-level and object-level F$_1$-scores for different numbers of training epochs. The third column shows the scores obtained with cross validation (i.e. the epoch is optimized) and the last column shows the rates corresponding to training for 15 epochs without any further optimization.

Considering these results, we notice that the single-class settings (SC, SE) do not show any useful results. In case of elliptical shapes (SE), at least the best configuration exhibits acceptable outcomes, however, GAN training is highly unstable in this scenario. 
In case of the multi-class settings (MC, ME), we notice a more stable behavior, as in each repetition good scores are obtained after few training epochs. Mean pixel-level F$_1$-scores of 0.63 (MC) and 0.62 (ME) as well as mean object-level F-scores of 0.74 (MC and ME) are achieved. Convergence is obtained approximately after six epochs for both settings. We notice slightly higher precision than recall, especially on object-level. A further optimization of the number the training epoch does not show a high influence.

The baseline results of the supervised approach are provided in Fig.~\ref{fig:baseline}. We notice that the break-even point of the supervised approach is reached with approximately four fully-annotated training WSIs corresponding to roughly 500 annotated glomeruli. Considering the object-level scenario, the proposed method exhibits even better performances (comparable with the supervised method trained on eight WSIs).

\begin{figure}[bt]
	\includegraphics[width=\linewidth]{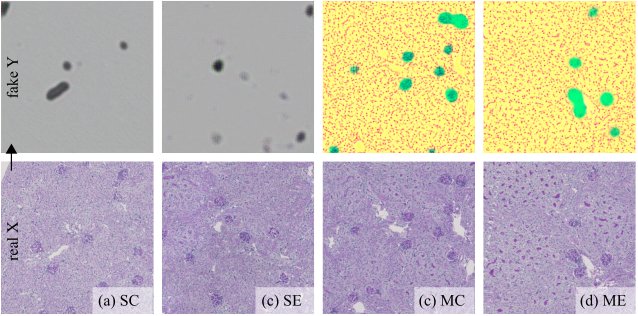}
	\caption{Qualitative results of the image translation process for the four different ettings (a) -- (d).}
	\label{fig:resQual}
\end{figure}

Example output of the image-to-image translation process is provided in Fig.~\ref{fig:resQual}. With the single-class setting ((a)--(b)), we notice a tendency to segment vessel structures instead of the target objects. This is not the case if making use of the multi-class settings ((c)--(d)).
%
%Qualitative segmentation results obtained with the MC setting are provided in Fig.~\ref{fig:resQualSeg}.
%We notice that especially small objects are often not detected partly leading to weak object-level F$_1$-scores.
%Additionally, we notice that in some cases only a small part of the object is segmented explaining the partially weak pixel-level scores.

%\begin{figure}
%	\centering
%	\subfloat[C]{\includegraphics[width=0.32\linewidth]{images/overall2_C_.pdf}}
%	\subfloat[E]{\includegraphics[width=0.32\linewidth]{images/overall2_E_.pdf}}
%	\subfloat[M]{\includegraphics[width=0.32\linewidth]{images/overall2_M_.pdf}}
%	\subfloat[ME]{\includegraphics[width=0.32\linewidth]{images/overall2_ME_.pdf}}
%	\caption{...}
%	\label{fig:plotOverall}
%\end{figure}

\section{Discussion}
In this work, we investigated a concept of fully-unsupervised learning for segmentation applications by making use of a GAN in combination with simulated annotation data.

We obtained highly divergent results for the four different settings.
One substantial finding is that a simulation of the annotations of the objects-of-interest only (referred to as single-class scenario) is not sufficient to obtain proper segmentations of the glomeruli in the investigated unpaired image-to-image translation scenario.
In the majority of attempts, an unwanted translation between the image and the label domain is observed.
A major problem here is that a translation from the label to the image domain cannot be performed which complicates the GAN training. The generator $G$ in this case has no chance to place the low-level objects (here the nuclei) in a way that the cycle consistency loss can become small as the position of the nuclei cannot be effectively derived from the annotation image. This behavior can also be seen in the example reconstructed images where nuclei cannot be clearly detected~(Fig.~\ref{fig:resQual}, third column).
In the multi-class scenarios with added simulated nuclei during GAN training, these objects are maintained during the training cycles. That means, the nuclei are segmented during translation to the label domain followed by a reconstruction of the nuclei based on the label domain in case of the inverse mapping.
%Interestingly, the texture of the glomeruli is still not reconstructed appropriately, which is likely due to the fact that the nuclei inside the glomeruli exhibit different properties (they are smaller and more elliptic) and are not perfectly segmented. However, this obviously does not strongly influence the final segmentation performance. Due to the small overall area of the glomeruli, the reconstruction error does not strongly influence the network's overall effectiveness.

A further interesting finding is that the distribution of the shapes of the simulated objects does not have a major impact on final segmentation performance. We do not consider the single-class scenarios here as they showed either completely wrong or highly unstable performance. The multi-class scenarios show similar performances for the setting based on circles and for the setting based on ellipses.

Considering the multi-class settings MC and ME, we assess the obtained segmentation performance as good and applicable for medical applications although the scores seem to be rather low. We need to mention here that this is on the one hand due to the fact that small objects are often not identified as glomeruli in the ground-truth but are detected by our approach. On the other hand, there are also small objects which are in the ground-truth but are not detected. Anyway, these objects are neglected by the medical experts and are thereby excluded from further analysis. %We do not consider a changed evaluation setting, in order to be able to compare the final scores.
%Often we also notice that only a small part of the objects is segmented (which is also indicated by low pixel-level scores compared to the high object-level scores). In order to improve the pixel-level behavior, a further iterative training approach could be applied as post-processing~\cite{myKhoreva17a} which could also be combined with higher image resolutions.

%For the proposed fully-unsupervised method, we did not consider an application to high resolution data so far, because the discriminator networks need large context in order to appropriately estimate whether an annotation is realistic of not. Therefore, if the image resolution is increased, the network architectures would also need to be changed accordingly. However, also from functional's perspective, a segmentation on the considered resolution is already highly useful.

A comparison with a state-of-the-art supervised approach showed that the novel method is highly competitive. Especially the detection performance (indicated by the object-level F$_1$-scores) is outperformed by the supervised technique only if training is performed with a large amount of annotated data (specifically with eight WSIs corresponding to approx. 1000 single objects).
Due to the stable training process, a "slightly-supervised" optimization of the training epoch is not required as the results are only marginally improved (Fig.~\ref{fig:res}, column 3 vs. column 4).

The most notable advantage, however, does not consist in high scores, but in a very high flexibility. The method can be easily adapted e.g. to other stains without a need for collecting novel annotated training data. Other applications can be handled as well by changing the simulation model. 
%Considering other application scenarios, a requirement is that the annotation data shows regularity and can be modeled effectively. If the objects-of-interest exhibit arbitrary shape, a conversion will not be possible based on unpaired data because a discriminator will not be able to distinguish between real and fake annotations.
%However, many segmentation tasks in biomedicine show such regular patterns and are thereby ideally suited for being processed with the proposed technique. 

%Questions:
%
% work in principle?
% how accurate virtual annotation do we need? (Object vs. pixel DSCs)
% effect of low-level information?
% outlook for other applications?
%

To conclude, we proposed and investigated a concept of fully-unsupervised learning for segmentation applications by making use of a GAN trained with real images and simulated annotations.
The experimental results, in general highly promising, indicate that it is not crucial to accurately model the underlying shape as long as a good approximation is available. This is a highly relevant finding as the shapes of the objects-of-interest are often too complex to be modeled accurately.
It is clearly more relevant to support the GAN to fulfill the cycle consistency criterion. Adding additional information to the label domain proved to be an effective way to facilitate the unpaired training process.
%To apply the proposed strategy to other application scenarios, the challenge consists in modeling the annotations and, even further, to model low-level structure to support the cyclic training process.
A comparison with a state-of-the-art supervised segmentation approach shows that the novel method is only outperformed if a large amount of labeled training data is available.

%\subsubsection*{Acknowledgement}
%*** Anonymous *** % TODO replace

%TODOs:
% boxplot instead of plots with std and min/max
% rates after all epochs
% rates after optimized epochs
% best rates

\bibliography{/home/staff/gadermayr/bibtex/my,/home/staff/gadermayr/bibtex/eigene}
\bibliographystyle{IEEEbib}

\end{document}